# Technical Note:

# Bias and the Quantification of Stability

Peter Turney
Knowledge Systems Laboratory
Institute for Information Technology
National Research Council Canada
Ottawa, Ontario, Canada
K1A 0R6

613-993-8564
peter@ai.iit.nrc.ca

**Abstract**

Research on bias in machine learning algorithms has generally been concerned with the impact of bias on predictive accuracy. We believe that there are other factors that should also play a role in the evaluation of bias. One such factor is the stability of the algorithm; in other words, the repeatability of the results. If we obtain two sets of data from the same phenomenon, with the same underlying probability distribution, then we would like our learning algorithm to induce approximately the same concepts from both sets of data. This paper introduces a method for quantifying stability, based on a measure of the agreement between concepts. We also discuss the relationships among stability, predictive accuracy, and bias.

# 1 Introduction

A significant area of research in machine learning involves empirical tests of algorithms that learn to classify. The most commonly used criterion for evaluation of classification algorithms is predictive accuracy. Other criteria are cost (e.g., the cost of acquiring data or the cost of classification errors) and complexity (e.g., the computational complexity of the algorithm or the syntactic complexity of the induced rules). We suggest another criterion: the *stability* of the algorithm.

The stability of a classification algorithm is the degree to which it generates repeatable results, given different batches of data from the same process. In our work with industrial applications of decision tree induction algorithms (Famili & Turney, 1991), we have discovered the importance of stability. We have used decision tree induction to generate rules that can predict low yield in a manufacturing process. The rules are used by process engineers to help them understand the causes of low yield. The engineers frequently have good reasons for believing that the causes of low yield are relatively constant over time. Therefore the engineers are disturbed when different batches of data from the same process result in radically different decision trees. The engineers lose confidence in the decision trees, even when we can demonstrate that the trees have high predictive accuracy.

A classification algorithm learns a concept from a set of training data. That concept is represented either explicitly (e.g., as a decision tree or a set of rules) or implicitly (e.g., as a set of stored instances, in the case of instance-based learning). To measure stability, we first need a measure of the similarity between two induced concepts. A syntactic measure of similarity (e.g., the percentage of overlap in the attributes used in two different decision trees) is likely to be *ad hoc* and specific to a particular concept representation. We use a semantic measure of similarity called *agreement* (Schaffer, 1992). Schaffer (1992) introduced the idea of agreement in his analysis of bias, but we use agreement here to analyze stability.

Section 2 presents a formal definition of stability, based on agreement. An empirical method for estimating stability is also presented. Stability is introduced here as a new criterion for evaluating biases. Section 3 discusses methods for improving stability. Section 4 summarizes the paper and lists some open questions.

# 2 Definitions and Theorems

In this section, we define and discuss predictive accuracy, agreement, stability, and bias.

## 2.1 Predictive Accuracy

The following notation is adapted from Schaffer (1992). Let $A$ be a finite or infinite set of attribute vectors and $C$ a finite set of classes. By a concept, $f$, we mean any function from $A$ to $C$. Let $D_{A \times C}$ be a probability distribution on $A \times C$. Let $t$ be $n$ samples from $A \times C$, where each sample is selected identically and independently from the distribution $D_{A \times C}$ (i.e., *iid* from $D_{A \times C}$). Given $A$ and $C$, let $F$ be the set of all concepts and let $T$ be the set of all training sets of size $n$, for some fixed $n$. A learning algorithm defines a function $L$ from $T$ to $F$. That is, a learning algorithm $L$ takes a training set $t$ in $T$ as input and generates a concept $f$ in $F$ as output.

The definition of predictive accuracy is familiar: we train the learner $L$ with data sampled from a distribution $D_{A \times C}$ and then test it with new data from the same distribution. Predictive accuracy is also known as generalization accuracy or testing set accuracy (to distinguish it from accuracy on the training set). In this paper, when we speak of accuracy, we mean predictive accuracy, not training set accuracy.

Let $t$ be selected with distribution $D_{A \times C}$. Let $f_{L(t)}$ be the concept learned by $L$ when given $t$ as input, $L(t) = f_{L(t)}$. Let $(a,c)$ be a random variable in $A \times C$ with $D_{A \times C}$.

**Definition 1:** The *predictive accuracy* of $L$ is defined to be $P_{D_{A \times C}}(f_{L(t)}(a) = c)$, the probability that $L$, when trained on a sample $t$ of length $n$, will correctly classify a new observation $(a,c)$. We use $\mathrm{acc}(L)$ to denote the predictive accuracy of $L$, given $D_{A \times C}$.

Definition 1 is a formal expression of the standard notion of predictive accuracy.

## 2.2 Agreement

The following definition of agreement is intended to capture the intuitive notion of similarity. The definition requires some explanation. In philosophical logic, there is a distinction between the *intension* and *extension* of a predicate. The extension of a predicate is the set of all things in the world for which the predicate is true. For example, the extension of the word "person" is the set of all people. The intension of a predicate is the denotation (meaning) of the predicate. [1]

The classical illustration of the difference between extension and intension is a story about a philosopher who defined "person" as "featherless biped". These phrases have the same extension (the set of all people), but different intensions (different meanings). According to the story, one of the philosopher's pupils obtained a chicken, plucked the chicken, and brought it back to her teacher. "Here is a person," she said.

Extension is relatively clear, but intension is harder to grasp. There is a suggestion, which can be traced back to Carnap (1947), and possibly earlier, that the intension of a predicate is its extension in all possible worlds. For example, "person" and "featherless biped" have the same extension in the real world, but there are possible worlds in which they have different extensions, as the pupil demonstrated. This leads naturally to the idea of measuring the similarity of two predicates by generating samples from "possible worlds" and seeing whether the predicates agree on the samples.

Let $D_A$ be a probability distribution over attribute vectors and let $a$ be a random variable in $A$ with distribution $D_A$. The distribution $D_A$ does not need to be the marginal distribution defined by $D_{A \times C}$; $D_A$ may be completely unrelated to $D_{A \times C}$. Let $f_1, f_2 \in F$ be any two concepts.

**Definition 2:** The *agreement* of $f_1$ and $f_2$ is defined to be $P_{D_A}(f_1(a) = f_2(a))$, the probability that $f_1$ and $f_2$ assign the random variable $a$ to the same class. We use agree($f_1, f_2$) to denote the agreement of $f_1$ and $f_2$, given $D_A$.

Definition 2 is from Schaffer (1992), where the concept of agreement was first introduced. Agreement is a measure of overlap in intension.

One might suppose that $D_A$ in Definition 2 should be defined as the marginal probability distribution given $D_{A \times C}$, which we could estimate from the training data. We prefer to define $D_A$ to be a uniform distribution over the set of all attribute vectors. The intention is to eliminate the statistical relationships between the attributes that are implicit in the distribution $D_{A \times C}$. The philosopher who defined "person" as "featherless biped" did so because, in his experience (given the training distribution $D_{A \times C}$), the attribute "person" is strongly correlated with the attribute "featherless biped". The pupil demonstrated the difference in intension of these attributes by creating a sample from a new distribution (the distribution $D_A$), where the correlation no longer exists. [2]

The following theorem shows an important property of agreement:

**Theorem 1:** Suppose that our attribute vectors are boolean and the concepts $f_1$ and $f_2$ are formulas in propositional calculus. Let us assume that there are $n$ boolean attributes and $A$ is the set of all possible attribute vectors, so $A = \{0, 1\}^n$ and $C = \{0, 1\}$. Suppose $D_A$ assigns a non-zero probability to every attribute vector in $A$. Then agree($f_1, f_2$) = 1.0 if and only if $f_1$ and $f_2$ are materially equivalent, $f_1 \equiv f_2$. [3]

*Proof:* This follows from the fact that agree($f_1, f_2$) = 1.0 if and only if $f_1$ and $f_2$ have the same truth-tables. By the semantic completeness and consistency of propositional calculus, $f_1 \equiv f_2$ if and only if $f_1$ and $f_2$ have the same truth-tables. □

For example, this theorem applies when $D_A$ is a uniform distribution. When $D_A$ is a uniform distribution, agree($f_1, f_2$) is the percentage overlap in the truth-tables of $f_1$ and $f_2$.

It is possible to determine the number of samples from $D_A$ that are required for a good estimate of agreement:

**Theorem 2:** Let us estimate the agreement of $f_1$ and $f_2$ by the average agreement, given $n$ samples from $D_A$. In the worst case, for *any* $D_A$, the standard deviation of the estimated agreement is less than or equal to $0.5/\sqrt{n}$.

*Proof:* We can consider $(f_1(a) = f_2(a))$ to be a random boolean function (1 if $f_1(a)$ and $f_2(a)$ are equal, 0 otherwise) of the random sample $a$. By definition, $(f_1(a) = f_2(a))$ is a sample from a Bernoulli distribution. There is a certain probability $p$ that $(f_1(a) = f_2(a))$ will be 1. We say that $(f_1(a) = f_2(a))$ is a sample from a Bernoulli($p$) distribution. Let $x_1, \ldots, x_n$ be $n$ samples from a Bernoulli($p$) distribution. Consider the average of $x_1, \ldots, x_n$. This average has a mean of $p$ and a standard deviation of $\sqrt{p(1-p)}/\sqrt{n}$ (Fraser, 1976). The worst case (largest standard deviation) is $p = 0.5$, where the standard deviation is $0.5/\sqrt{n}$. □

If we set $n = 10{,}000$, then the standard deviation is 0.005 (in the worst case), or 0.5%, which seems acceptably low. Suppose $A = \{0, 1\}^s$, so $|A| = 2^s$. The truth-tables for $f_1$ and $f_2$ would have $2^s$ rows. As $s$ grows, a precise calculation of the agreement of $f_1$ and $f_2$ quickly becomes infeasible. For example, with $s = 30$, we would need to look at $2^{30} \approx 10^9$ boolean vectors. Theorem 2 shows that we can get a good estimate of agreement by looking at only $10^4$ boolean vectors.

Intension (meaning, denotation) and extension (reference) are both *semantic* notions. Agreement is a semantic measure of intensional similarity. One might ask why we do not use a *syntactic* measure of similarity, since syntactic measures are more familiar, perhaps more intuitive, and easier to compute than the semantic measure introduced here. For example, we could use Levenshtein edit distance (Levenshtein, 1966; Honavar, 1992) to measure the syntactic similarity of two decision trees. [4] There are several problems with syntactic measures of similarity. First, they tend to be *ad hoc*. Second, they are dependent on the chosen representation. This means that we would need to develop a different syntactic similarity measure for each different representation that we consider. It also means that we would not be able to compare stability across distinct representations. For example, we could not compare the stability of a decision tree induction algorithm with the stability of a neural network algorithm, if we based our definition of stability on a syntactic measure of similarity. Third, a syntactic similarity measure would count logically equivalent representations as different. For example, suppose we have two superficially different decision trees. Suppose we translate the decision trees into disjunctive normal form expressions in propositional calculus and it turns out that the two expressions are logically equivalent. We know by Theorem 1 that the agreement of the two decision trees is 1. However, the Levenshtein edit distance between the two trees could be quite large. Unlike a syntactic similarity measure, agreement is not sensitive to superficial differences in representations. This is a virtue of agreement, since we should not be concerned with superficial differences in the expression of a concept. Our concern should be with differences in concept *meaning* (intension). Sometimes it is not readily apparent that two different representations of a concept are in fact logically equivalent. However, by empirically estimating their agreement, we can discover their logical equivalence (more precisely, their logical similarity).

## 2.3 Stability

When a learner $L$ is trained on two sets of data $t_1$ and $t_2$ that are sampled from the same distribution $D_{A \times C}$, we would like the learned concepts $f_{L(t_1)}$ and $f_{L(t_2)}$ to have approximately the same intension. Even when both concepts $f_{L(t_1)}$ and $f_{L(t_2)}$ have high predictive accuracy, we find it disturbing when the concepts have radically different intensions. Definition 3 is an attempt to capture this idea.

Let $t_1$ and $t_2$ be two distinct *iid* sequences of samples of length $n$, selected with distribution $D_{A \times C}$. Let $f_{L(t_1)}$ and $f_{L(t_2)}$ be random variables that represent the concepts learned by $L$, given the training sequences $t_1$ and $t_2$, respectively.

**Definition 3:** The *stability* of $L$ is defined to be the expected agreement of $f_{L(t_1)}$ and $f_{L(t_2)}$, $E_{D_{A \times C}}(\text{agree}(f_{L(t_1)}, f_{L(t_2)}))$. We use stable($L$) to denote the stability of $L$, given $D_{A \times C}$ and $D_A$. Combining Definitions 2 and 3, the stability of $L$ is:

$$E_{D_{A \times C}}(P_{D_A}(f_{L(t_1)}(a) = f_{L(t_2)}(a))) \tag{1}$$

Definition 3 is new.

Again, $D_A$ may be completely unrelated to $D_{A \times C}$. Suppose that the attributes $a_1$ and $a_2$ are highly correlated, given the distribution $D_{A \times C}$. Suppose that we use C4.5 (Quinlan, 1992) to learn decision trees on two sets of data sampled from $D_{A \times C}$. When building a decision tree, C4.5 selects attributes using the information gain ratio. Since $a_1$ and $a_2$ are highly correlated, C4.5 may consider them to be equally acceptable, according to their information gain ratios. It may happen that $a_1$ has a slightly higher information gain ratio than $a_2$ in the first training set, but $a_2$ has a higher ratio than $a_1$ in the second training set. Thus the first decision tree might use $a_1$ while the second decision tree uses $a_2$. Thus highly correlated attributes can be one source of instability. To detect this instability, we cannot compare the agreement of the two decision trees by defining $D_A$ as the marginal distribution given $D_{A \times C}$. We must use a distribution $D_A$ where the correlation between $a_1$ and $a_2$ has been eliminated.

We usually do not have direct knowledge of the distribution $D_{A \times C}$; typically we have a set of training data $t$, consisting of samples from an unknown distribution $D_{A \times C}$. We can estimate predictive accuracy and stability using a standard technique, $n$-fold cross-validation, with $n = 2$. We randomly split the training data into two equal-size subsets (nearly equal, if the training set has an odd size). We train the learner on one subset, then test it on the second subset. We then swap the subsets and repeat the process. To increase the reliability of the estimates for predictive accuracy and stability, we repeat the 2-fold cross-validation $m$ times. Figure 1 summarizes the method.

The parameters $m$ and $n$ in Figure 1 determine the quality of the estimates for predictive accuracy and stability. Larger values will yield better estimates, but they will also require more computer time. Theorem 2 gives some guidance in determining a suitable value for $n$. The following theorem gives some guidance for $m$.

**Theorem 3:** Let us estimate the stability of $L$ by the method given in Figure 1. In the worst case, for any $D_A$ and any $D_{A \times C}$, the standard deviation of the estimated stability is less than or equal to $0.5 / \sqrt{m}$.

*Proof:* In the method given in Figure 1, the expected agreement of $f_{L(t_1)}$ and $f_{L(t_2)}$ is estimated by $\text{stab}_i$, which is the average of $(f_{L(t_1)}(a) = f_{L(t_2)}(a))$ given $n$ samples from the distribution $D_A$, where $D_A$ could be, for example, a uniform distribution. Since $(f_{L(t_1)}(a) = f_{L(t_2)}(a))$ has a Bernoulli($p$) distribution, $n \cdot \text{stab}_i$ has a binomial($n, p$) distribution (Fraser, 1976). The stability of $L$, stable($L$), is estimated by $\overline{\text{stab}_i}$, the average of $\text{stab}_i$ for $i$ equal 1 to $m$. Therefore $n \cdot m \cdot \overline{\text{stab}_i}$ is the sum of $m$ random variables with binomial distributions, binomial($n, p_1$), …, binomial($n, p_m$). These $m$ binomial variables may have different probabilities $p_i$, due to variation in the $m$ random splits of the data. The range of values for $p_i$ depends on the distribution $D_{A \times C}$, which we assume is unknown. In the worst case, $p_i$ is itself a random variable with a Bernoulli(0.5) distribu-

    For $i = 1$ to $m$ do:
        Randomly split the training data $t$ into two equal-size subsets $t_1$ and $t_2$
        Run the learner $L$ on the data $t_1$, resulting in the concept $f_{L(t_1)}$
        Run the learner $L$ on the data $t_2$, resulting in the concept $f_{L(t_2)}$
        Let $acc_{i,1}$ be the accuracy of $f_{L(t_1)}$ on the data $t_2$
        Let $acc_{i,2}$ be the accuracy of $f_{L(t_2)}$ on the data $t_1$
        For $j = 1$ to $n$ do:
            Generate a random attribute vector $a$ from $D_A$
            Let $agr_j$ be 1 if $f_{L(t_1)}(a) = f_{L(t_2)}(a)$, 0 otherwise
        End for $j$
        Let $stab_i$ be the average of $agr_j$, for $j = 1$ to $n$
    End for $i$
Let the estimated predictive accuracy be the average of $acc_{i,1}$ and $acc_{i,2}$, for $i = 1$ to $m$
Let the estimated stability be the average of $stab_i$, for $i = 1$ to $m$

Figure 1. Pseudocode for empirical approximation of predictive accuracy and stability.

tion. In this case, $n \cdot stab_i$ is either 0 or $n$, so $stab_i$ is either 0 or 1. That is, $stab_i$ will also have a Bernoulli(0.5) distribution. Therefore the result of Theorem 2 applies and $\overline{stab_i}$ has a standard deviation less than or equal to $0.5/\sqrt{m}$. □

Note that, in the worst case, the value of $n$ is irrelevant in determining the standard deviation of stability. This would suggest that a good strategy would be to set $n$ to 1 and $m$ to 10,000 (for example). Unfortunately, the inner loop of the procedure in Figure 1 (for $j$ equal 1 to $n$) will usually be much more efficient to compute than the outer loop (for $i$ equal 1 to $m$). The outer loop is where the learner actually learns, while the inner loop merely involves the application of what has been learned. In general, the value of $n$ will have some impact on the standard deviation of stability. Thus concern for computational efficiency suggests that $n$ should be more than 1 and $m$ should be less than 10,000. The appropriate values of $m$ and $n$ must be chosen by trading off the accuracy of the estimates for predictive accuracy and stability, on the one hand, and the computational efficiency of the procedure in Figure 1, on the other hand.

A stable learning algorithm can be used to detect change in a stochastic process.

Suppose we continuously collect batches of data from some ongoing data generating process (e.g., a manufacturing process — more generally, samples from a probability distribution). As each batch of data arrives, we give it to our learning algorithm to analyze. If our learning algorithm is stable, a small change in the extension of the target concept (due to random fluctuations in the batches of data) will result in a small change in the intension of the induced concept. A large change in the intension of the induced concept implies a large change in the extension of the target concept. Thus we can use the intension of the induced concept to monitor variation in the extension of the target concept. For example, in manufacturing process control, a process engineer can detect changes in the manufacturing process by monitoring the rules that are induced by the learning algorithm, over successive batches of data. On the other hand, if the learning algorithm is unstable, variation in the intension of the induced concept is not proportional to variation in the extension of the target concept. Therefore an unstable learning algorithm cannot be used to monitor change in the data generating process. This is the central motivation for desiring stable learning algorithms.

In Section 3, we examine ways to improve the stability of a learner. One way to increase stability is to increase the strength of bias. Therefore, in Section 2.4, we discuss bias, in preparation for our discussion of ways to improve stability.

## 2.4  Bias

Utgoff (1986) describes biases along the dimensions of *strength* and *correctness*:

1. A *strong* bias is one that focuses the concept learner on a relatively small number of hypotheses. Conversely, a *weak* bias is one that allows the concept learner to consider a relatively large number of hypotheses.
2. A *correct* bias is one that allows the concept learner to select the target concept. Conversely, an *incorrect* bias is one that does not allow the concept learner to select the target concept.

Rendell (1986) distinguishes two types of bias, *exclusive* bias and *preferential* bias. A learning algorithm has an exclusive bias against a certain class of concepts when the algorithm does not even consider any of the concepts in the class. A less extreme form of bias is a preference for one class of concepts over another class.

Some researchers distinguish *representational* bias and *procedural* bias. Representational bias is typically a form of exclusive bias, since constraining the representation language means that certain concepts cannot be considered, since they cannot be expressed. Procedural bias is typically a form of preferential bias. For example, pruning in C4.5 is a procedural bias that results in a preference for smaller decision trees (Quinlan, 1992; Schaffer, 1993). The distinction between exclusive and preferential bias is based on the learner's behavior, while the distinction between representational and procedural bias is based on the learner's design.

Utgoff's (1986) definition of bias strength implicitly assumes an exclusive bias. A learner with a strong exclusive bias (equivalently, a strong representational bias) considers only a small class of concepts. The VC (Vapnik-Chervonenkis) dimension is a measure of the strength of an exclusive bias (Vapnik, 1982; Haussler, 1988). Schaffer (1992) introduced the definition of agreement (Definition 2 here) as a component in his definition of a measure of the strength of preferential bias. The idea behind Schaffer's definition is that a learner with a strong preferential bias toward $f_1$ over $f_2$ will only learn $f_2$ when $f_2$ is substantially more accurate than $f_1$.

The following is a variation on Schaffer's (1992) definition. Let $t$ be a random variable representing a training set of size $n$ sampled *iid* from the distribution $D_{A \times C}$ over $A \times C$. Let the learned concept $f_L$ be a random variable that depends on the value of the random variable $t$. The expectation $E_{D_{A \times C}}$ in the following definition is calculated with the random variable $f_L$ using the distribution $D_{A \times C}$.[5]

**Definition 4:** We say $L$ *prefers* $f_1$ to $f_2$ if $E_{D_{A \times C}}(\mathrm{agree}(f_1, f_L)) > E_{D_{A \times C}}(\mathrm{agree}(f_2, f_L))$.

That is, $L$ prefers $f_1$ to $f_2$ if we expect $L$ to learn a concept $f_L$ that agrees more with $f_1$ than with $f_2$. In this definition, the preference of $L$ depends on the distribution $D_{A \times C}$. This leads naturally to the idea that we can measure the strength of the learner's preference by varying the distribution.

Let $\delta(c_1, c_2)$ be defined as follows:

$$\delta(c_1, c_2) = \left\{ \begin{array}{l} 1 \text{ if } c_1 = c_2 \\ 0 \text{ if } c_1 \neq c_2 \end{array} \right\} \quad (2)$$

Let $f_1$ and $f_2$ be any two concepts. Let $p \in [0,1]$. Let $D'_A$ be any distribution on $A$, where $D'_A$ may be distinct from the distribution $D_A$ used to define agreement. $D'_A(a)$ is the probability of $a$ according to the distribution $D'_A$. We may define a family of distributions $D°((a,c) | D'_A, p, f_1, f_2)$ for $(a,c) \in A \times C$, given $D'_A$, $p$, $f_1$, and $f_2$, as follows:

$$D°((a,c) | D'_A, p, f_1, f_2) = (1-p) \cdot \delta(f_1(a), c) \cdot D'_A(a) + p \cdot \delta(f_2(a), c) \cdot D'_A(a) \quad (3)$$

Given a random observation $(a, c)$ sampled with distribution $D°((a,c) | D'_A, p, f_1, f_2)$,

the probability that $f_1$ correctly classifies $a$ is $p$, while the probability that $f_2$ correctly classifies $a$ is $1-p$, assuming $f_1$ and $f_2$ disagree on $a$. The accuracy of $f_1$ may be greater than $p$ and the accuracy of $f_2$ may be greater than $1-p$, since they may agree.

Let $t_p$ be a random variable representing a training set consisting of an *iid* sequence of $n$ samples from the distribution $D°((a,c)\,|\,D'_A, p, f_1, f_2)$.

**Definition 5:** If $L$ prefers $f_1$ to $f_2$ for training sets $t_{0.5}$, we say the learner is *biased* toward $f_1$ over $f_2$.

A training set $t_{0.5}$ provides, on average, equal evidence for both $f_1$ and $f_2$. If the learner has a preference when the evidence is ambiguous, $p = 0.5$, it is natural to call this preference a bias. We can extend this idea further to provide a measure of the strength of the learner's bias.

**Definition 6:** If $p > 0.5$ and $p$ is the smallest value such that $L$ prefers $f_1$ to $f_2$ for training sets $t_p$, we say the learner has a *preferential bias of strength $p$* toward $f_1$ over $f_2$.

Thus the strength of a preferential bias ranges from greater than 0.5 to less than or equal to 1.0. Speaking metaphorically, $p$ represents the "force" that the data must "exert" to overcome the bias of $L$. This definition readily lends itself to empirical measurement of bias strength (Schaffer, 1992). We can generate artificial data, using the distribution $D°((a,c)\,|\,D'_A, p, f_1, f_2)$, and we can vary the value of $p$ to discover the strength of the preferential bias of $L$.

A strong exclusive bias accelerates learning by reducing the learner's search space. A strong bias (either exclusive or preferential) also improves resistance to noise in the training data. Increasing bias can increase accuracy, if the bias pushes the learner towards more accurate concepts. However, increasing bias can also decrease accuracy, if the bias pushes the learner in the wrong direction (Schaffer, 1993). For example, if we restrict the representational power of the learner too severely, the target concept may lie outside of what the learner can represent. The difficulty is determining the right direction for bias.

Utgoff's (1986) definition of bias correctness implicitly assumes that predictive accuracy is the measure of the correctness of a bias. PAC (probably approximately correct) learning provides another measure of bias correctness (Haussler, 1988). We propose that the notion of bias correctness should be extended to include stability. More generally, criteria for the correctness of a bias include accuracy, stability, cost (e.g., the cost of acquiring data or the cost of classification errors), and complexity (e.g., the computational complexity of the algorithm or the syntactic complexity of the induced concepts).

## 3 Improving Stability

There are at least three ways to improve the stability of a learner. First, we can increase the size of the training data sample $t$. With larger samples, the learner is less likely to be influenced by random sampling effects. Second, we can increase the strength (either exclusive or preferential) of the bias. This increases stability by pushing the learner towards a given class of concepts. Third, we can have a learner that memorizes previously learned concepts. When we run the learner on the data $t_1$, it memorizes the resulting concept $f_{L(t_1)}$. When we run the learner on the data $t_2$, it generates a new concept $f_{L(t_2)}$, but it also considers the old concept $f_{L(t_1)}$. Given a certain desired level of accuracy and stability, the learner decides between $f_{L(t_1)}$ and $f_{L(t_2)}$ as the output concept for the data $t_2$.

Stability is similar to accuracy, in that the above three techniques will also affect the accuracy of the learner. However, stability is quite different from accuracy with regard to the second technique, since *any* increase in bias strength will cause an increase in stability. The direction of the bias does not matter. We can trivially maximize stability by making a 'learner' with a constant output. However, in most applications, we are not interested in stability alone, but stability in conjunction with accuracy. We do not claim that stability is always desirable. For example, in some situations, we may want to discover all concepts that are consistent with the training data (Murphy & Pazzani, 1994). Thus we may sometimes define a measure of bias correctness that is based on accuracy alone, but there would not be much interest in a measure based on stability alone.

If our measure of bias correctness incorporates both accuracy and stability, then the task of selecting a bias is easier than when the measure is based on accuracy alone. When accuracy is our only concern, the bias must be chosen to push the learner towards the target concept. When stability is important, a slight bias away from the target concept may have a net benefit, if the increase in stability is greater than the decrease in accuracy.

## 4 Discussion

In this paper we defined a measure of the stability of learning algorithms and we presented an empirical technique for quantifying stability. We argued that stability is a desirable property in learning algorithms. Stability is especially important to users who do not understand the inner working of a learning algorithm. Even when the algorithm generates rules that are easy to understand, the user views the algorithm as an oracle. It is difficult to trust an oracle, and it is especially difficult to trust an oracle that says something radically different each time you make a slight change to the wording of your question.

There are many questions about stability and its relation to accuracy and bias. How do distinct learning algorithms, such as neural networks and decision trees, compare with respect to stability? How can a learning algorithm adjust the strength and type of its bias to suit the data it faces, to optimize the correctness of its bias? When measuring the correctness of a bias, how can we combine the criteria of accuracy, stability, cost, and complexity? How do we handle the case where the set of classes, $C$, is infinite? That is, how do we measure stability when learning to fit a curve, rather than learning to classify? These are open questions.

## Acknowledgments

My thanks go to the following people: Michael Jankuľák and John Wong provided invaluable assistance in conducting experiments on the stability of C4.5. Cullen Schaffer discussed his own, related work and provided useful feedback. Malcolm Forster gave several helpful comments on an early version of this paper. Three anonymous referees of *Machine Learning* provided extensive, helpful comments and criticism, as did Diana Gordon and Marie desJardins.

## Notes

1. In other words, the extension of a predicate is that to which the predicate refers and the intension of a predicate is the meaning of the predicate. The technical concept of *intension* is introduced in this paper in an effort to capture an aspect of the informal, intuitive concept of *meaning*. For a philosophical justification of this effort, the reader may turn to Carnap (1947). Meaning is relevant here, because we are seeking a notion of similarity that is based on meaning.

2. To expose the differences between two concepts $f_1$ and $f_2$, we need a distribution $D_A$ that yields correlations that contrast with those given by the marginal distribution defined by $D_{A \times C}$. The uniform distribution is one possible choice for $D_A$. Another possibility would be to choose a distribution $D_A$ in which the correlations between pairs of attributes have the opposite signs of the correlations in the marginal distribution defined by $D_{A \times C}$. However, in general, this is not possible. Suppose we have three attribute vectors, such that $\text{correlation}(a_1, a_2) = 1$, $\text{correlation}(a_1, a_3) = -1$, and $\text{correlation}(a_2, a_3) = -1$. If we switch the signs of the correlations, we will get $\text{correlation}(a_1, a_2) = -1$, $\text{correlation}(a_1, a_3) = 1$, and $\text{correlation}(a_2, a_3) = 1$, but this is inconsistent. If $\text{correlation}(a_1, a_3) = 1$ and $\text{correlation}(a_2, a_3) = 1$, then it

  necessarily follows that correlation$(a_1, a_2) = 1$.

3. The English statement, "*A* if and only if *B*," is expressed by logicians as, "*A* is materially equivalent to *B*," which is written symbolically as $A \equiv B$.

4. The Levenshtein edit distance measures the distance between two structures (e.g., strings, trees) by the number of edit operations (e.g., deletions, insertions, reversals) required to transform one structure into the other.

5. Note that, in the definitions in this section, $f_1$ and $f_2$ are two fixed concepts. This is different from the previous section, where $f_{L(t_1)}$ and $f_{L(t_2)}$ are random variables.